\documentclass[runningheads]{llncs}

\usepackage{accv}

\usepackage{accvabbrv}

\usepackage{graphicx}
\usepackage{booktabs}

\usepackage[accsupp]{axessibility}  

\usepackage{subcaption}
\usepackage{bm}
\usepackage{multirow}
\usepackage{makecell}
\usepackage{pifont}

\newcommand{\xmark}{\ding{55}}

\usepackage{hyperref}

\usepackage{orcidlink}

\begin{document}

\title{Enhancing 3D Human Pose Estimation with Bone Length Adjustment} 


\author{Chih-Hsiang Hsu\inst{1}\orcidlink{0009-0000-4401-1324} \and
Jyh-Shing Roger Jang\inst{1}\orcidlink{0000-0002-7319-9095}}

\authorrunning{C. Hsu and J. Jang}

\institute{Graduate Institute of Networking and Multimedia, National Taiwan University, Taiwan \\
\email{r11944034@ntu.edu.tw}}

\maketitle

\begin{abstract}

Current approaches in 3D human pose estimation primarily focus on regressing 3D joint locations, often neglecting critical physical constraints such as bone length consistency and body symmetry. This work introduces a recurrent neural network architecture designed to capture holistic information across entire video sequences, enabling accurate prediction of bone lengths. To enhance training effectiveness, we propose a novel augmentation strategy using synthetic bone lengths that adhere to physical constraints. Moreover, we present a bone length adjustment method that preserves bone orientations while substituting bone lengths with predicted values. Our results demonstrate that existing 3D human pose estimation models can be significantly enhanced through this adjustment process. Furthermore, we fine-tune human pose estimation models using inferred bone lengths, observing notable improvements. Our bone length prediction model surpasses the previous best results, and our adjustment and fine-tuning method enhance performance across several metrics on the Human3.6M dataset. The code is available at:
\url{https://github.com/hs1ang-hsu/BLAPose}

\end{abstract}

\section{Introduction}
\label{sec:intro}

3D human pose estimation aims to localize the 3D positions of human joints from monocular images or videos, holding significant implications for applications such as human-computer interaction, sports analysis, and medical diagnostics, due to its capacity to capture human motion. Presently, two-stage approaches dominate in 3D human pose estimation. These methods initially detect 2D keypoints from input images or videos and subsequently lift these 2D keypoints into 3D space, which is known as the 2D-to-3D lifting task.

The 2D-to-3D lifting task faces inherent challenges due to depth ambiguity: multiple 3D poses can project to the same 2D keypoints. A simple example is the flipping ambiguity described by \cite{Sminchisescu2003kinematic}, where finite possible 3D poses arise by flipping each bone forwards or backwards. This ambiguity intensifies when bone lengths are unknown or when 2D keypoints are inaccurate. Recognizing the importance of temporal information in resolving depth ambiguity, recent studies have employed recurrent neural networks (RNNs) \cite{hossain2018exploiting,lee2018propagating}, temporal convolutions \cite{pavllo20193d:videopose,chen2021anatomy}, and transformer-based models \cite{zheng20213d:poseformer,li2022mhformer,zhang2022mixste} to extract temporal features.

Despite recent advancements, many existing methods overlook the natural structure of human poses. Studies have shown that focusing solely on minimizing per-joint errors independently neglects overall pose coherence. Addressing this issue, bone-based representations have been proposed \cite{sun2017compositional}. Chen \etal \cite{chen2021anatomy} introduced a method to decompose human poses into bone lengths and directions, simplifying the pose estimation task. However, integrating physical constraints such as bone length consistency and body symmetry remains a challenge, with significant bone length errors observed in existing works.

Inspired by previous work \cite{chen2021anatomy}, we propose RNN-based models to predict bone lengths. Our models leverage global information from all frames of a video, rather than short sequences. To enhance training effectiveness, we introduce a novel training time augmentation method using synthetic bone lengths generated by the statistical body shape model called SMPL \cite{Loper2015SMPL}. This augmentation ensures that predicted bone lengths respect symmetry constraints and adhere to natural human body proportions. Unlike the method in \cite{chen2021anatomy} that randomly adjust bone lengths, potentially distorting body proportions, our method prioritizes realistic and accurate predictions.

We propose a novel adjustment method to enhance current state-of-the-art 2D-to-3D lifting models. This method preserves bone directions while replacing bone lengths, ensuring that the adjusted poses maintain anatomical correctness and achieve more precise joint positions. Finally, we fine-tune existing 2D-to-3D lifting models using bone length information. This fine-tuning process further improves model performance and can be applied to any 2D-to-3D lifting model, demonstrating its versatility and effectiveness.

The contributions of this work are threefold:

\begin{itemize}
    \item We propose a bone length prediction model that effectively utilizes global information and synthetic augmentation to predict accurate bone lengths.
    
    \item We introduce a bone length adjustment method that enhances existing 2D-to-3D lifting models, ensuring realistic body shapes and accurate joint positions.

    \item We demonstrate the efficacy of fine-tuning existing models with predicted bone lengths, thereby improving their performance in 3D human pose estimation tasks.
\end{itemize}

\section{Related Work}

3D human pose estimation using deep learning methods can be categorized into two main approaches: the end-to-end approach and the two-stage approach.

The end-to-end approach predicts 3D human poses directly from RGB images. These methods \cite{kanazawa2018end,kocabas2020vibe,wei2022capturing} heavily rely on parametric 3D human shape models, such as SMPL \cite{Loper2015SMPL}. They estimate parameterized pose, shape, and translation, which are then decoded via the SMPL model to obtain the human pose and shape.

In contrast, the two-stage approach first detects 2D positions of joints in an image, known as 2D keypoints. Techniques such as convolutional neural networks (CNNs) generate heatmaps that indicate the probability of joint locations \cite{Chen2018cascaded:CPN,Sun2019deep:hrnet}. In the second stage, these 2D keypoints are used to estimate the 3D pose of the human figure, which is called 2D-to-3D lifting task. Currently, the two-stage approach tends to be more accurate than the end-to-end approach. End-to-end methods struggle with a lack of diversity in video data, especially variations in background, due to the complex requirements of 3D pose datasets, such as motion capture systems and high-speed cameras. This issue is mitigated in the two-stage approach since 2D keypoints can be manually labeled, and modern 2D keypoint detectors achieve high precision.

The 2D-to-3D lifting task is an ill-posed problem that involves predicting the additional depth dimension. Recent works have utilized temporal information by predicting the human pose of the central frame from a sequence of 2D keypoints in a video. The context information provides clues about the movement of the target, making the predicted poses more robust to noise. Hossain and Little \cite{hossain2018exploiting} designed an LSTM sequence-to-sequence model to obtain temporally consistent 3D poses. Pavllo \etal \cite{pavllo20193d:videopose} proposed a model based on dilated temporal convolution to capture long-term information and reduce computational overhead compared to LSTM models. Chen \etal \cite{chen2021anatomy} demonstrated that dividing the task into bone length and bone direction prediction yields better results.


Since pictorial information is often lost in the 2D-to-3D lifting task, some works have introduced image clues as input, such as the root depths of targets \cite{liu2022explicit,zhan2022ray3d,cheng2021graph} and occlusion scores of keypoints \cite{chen2021anatomy,liu2022explicit}. Zhao \etal \cite{zhao2024single} exploited intermediate visual representations of 2D pose detectors, achieving promising results even when predicting from a single image.

However, the consistency and accuracy of bone lengths have not been well considered in these approaches. Our research demonstrate that these models can be further improved by adjusting bone lengths, ensuring more accurate and realistic predictions.

\section{Methods}

Our work consists of two main components: bone length prediction and bone length adjustment. Section \ref{sect:length aug} details our bone length augmentation method, Section \ref{sect:length model} discusses our model design for predicting bone lengths, and Section \ref{sect:length adjustment} explains how we apply our bone length prediction model to enhance 2D-to-3D lifting models.

\subsection{Bone Length Augmentation}\label{sect:length aug}

\subsubsection{Augmentation.}

\begin{figure}
    \centering
    \begin{subfigure}[b]{0.2\textwidth}
        \centering
        \includegraphics[height=40mm]{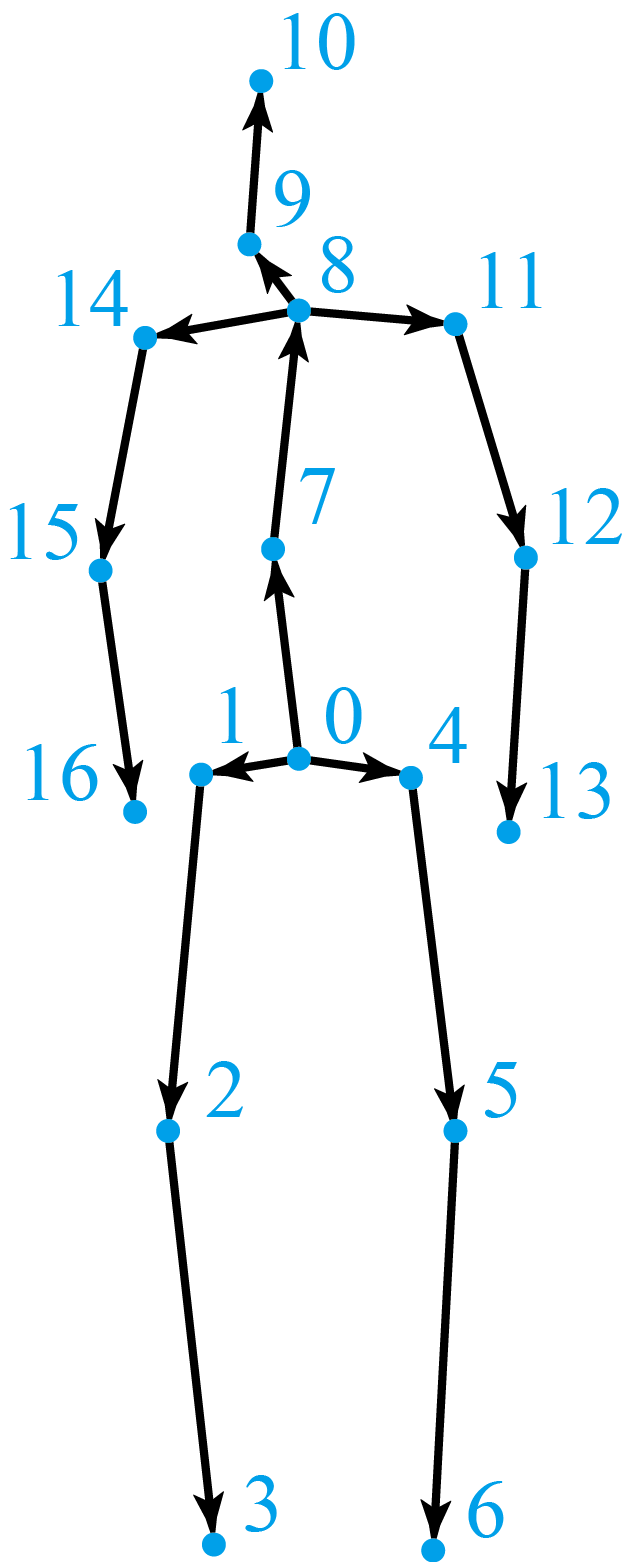}
        \caption{}
        \label{fig:skeleton}
    \end{subfigure}
    \hfill
    \begin{subfigure}[b]{0.7\textwidth}
        \centering
        \includegraphics[height=35mm]{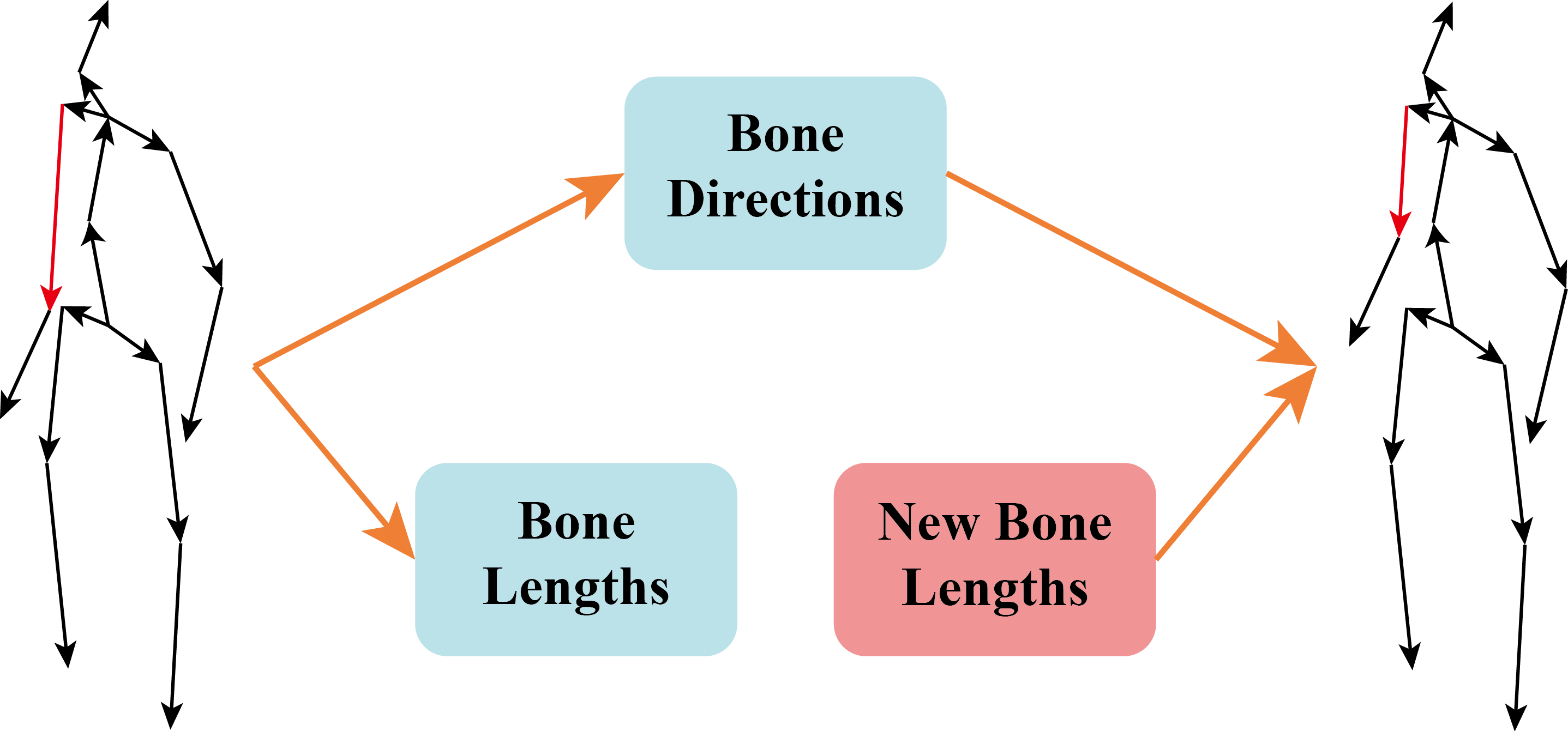}
        \caption{}
        \label{fig:replacement}
    \end{subfigure}

    \caption{
    (a) The representation of a human pose with joint labels. 
    (b) The overview of bone length replacement, which involves decomposing the pose into bone directions and bone lengths, and then substituting the original bone lengths with new ones.
    }
\end{figure}

We represent a human pose $P = \left[ \bm{p}_0 \cdots \bm{p}_{J-1} \right]^T \in \mathbb{R}^{J \times 3}$ with $J$ 3D joint positions as a tree structure, as shown in Figure \ref{fig:skeleton}. The root joint is positioned on the pelvis and labeled as joint $0$. For each joint $\bm{p}_i \in \mathbb{R}^3$, its parent is defined as the joint closer to the root (\eg, joint 0 is the parent of joint 1). The pose can be decomposed into bone lengths $L = \left[ l_1 \cdots l_{J-1} \right]^T \in \mathbb{R}^{(J-1) \times 1}$ and bone directions $D = \left[ \bm{d}_1 \cdots \bm{d}_{J-1} \right]^T \in \mathbb{R}^{(J-1) \times 3}$ using the following equations:

\begin{equation}
    \begin{split}
        l_i &= \Vert \bm{p}_i - \bm{p}_{\text{parent}(i)} \Vert_2,\quad i=1,\dots,J-1 \\
        \bm{d}_i &= \frac{\bm{p}_i - \bm{p}_{\text{parent}(i)}}{l_i},\quad i=1,\dots,J-1
    \end{split}
    \label{eq:pose decomposition}
\end{equation}

Here, vertices (joints) are labeled from $0$ to $J-1$ and the edges (bones) are labeled from $1$ to $J-1$. Given $L$ and $D$, the original pose $P$ can be reconstructed.

In the augmentation process, we first decompose a pose $P$ into bone lengths $L$ and bone directions $D$. We then use new bone lengths $L' = \left[ l'_1 \cdots l'_{J-1} \right]^T$ and the original bone directions $D$ to reconstruct a new pose $\tilde{P} = \left[ \tilde{\bm{p}}_1 \cdots \tilde{\bm{p}}_{J-1} \right]^T$. The bone length replacement process is illustrated in Figure \ref{fig:replacement}. A random shift $\bm{s} \in \mathbb{R}^3$ is added to the poses to enhance the augmentation. The final result is the augmented pose $P' = \left[ \bm{p}'_1 \cdots \bm{p}'_{J-1} \right]^T$.

\begin{equation}
    \begin{split}
        \bm{s} &\sim \mathcal{N}(\mu=0,\sigma=0.5) \\
        \bm{p}'_i &= \tilde{\bm{p}}_i + \bm{s},\quad i=1,\dots,J-1
    \end{split}
    \label{eq:random shift}
\end{equation}

For a sequence of poses, the same random shift should be applied to preserve smoothness in the trajectory. Since we use 2D keypoints as the model input, we project the pose $P'$ onto the 2D camera plane considering the camera intrinsic matrix (focal length and principal point), and both radial and tangential nonlinear lens distortion. Our model learns to predict $L'$ from the projected 2D keypoints. The key to the augmentation process is generating reasonable bone lengths $L'$.

\subsubsection{Random Bone Lengths.}

\begin{figure}
\centering
\includegraphics[height=50mm]{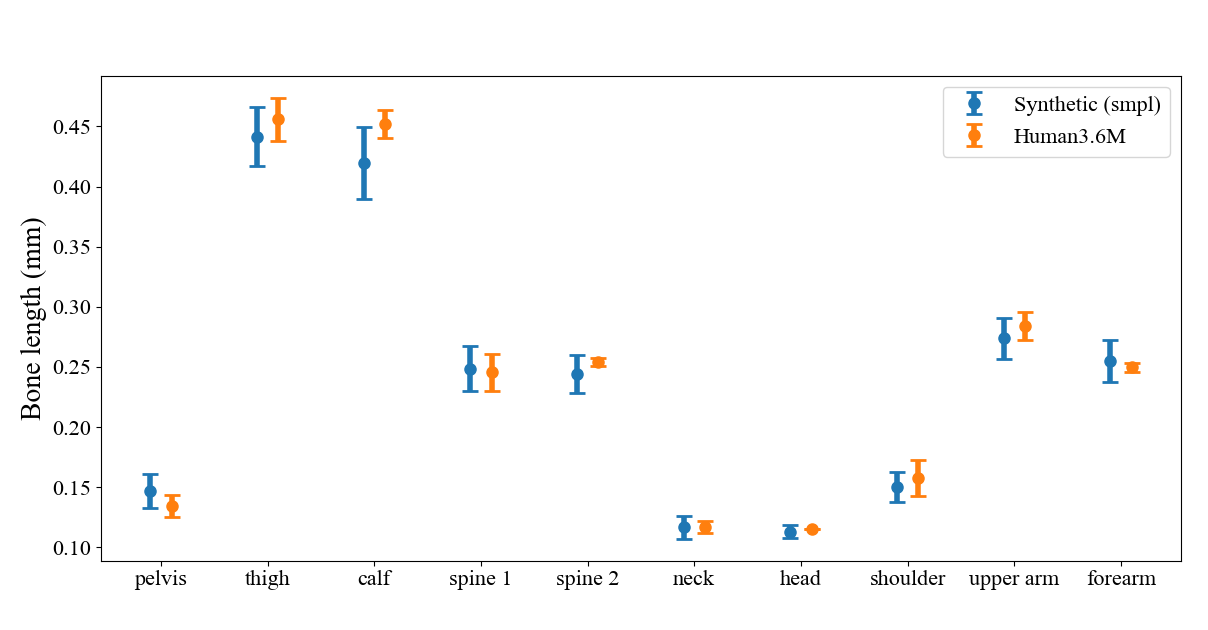}
\caption{This error bar plot shows the means and the standard deviations of bone lengths in the Human3.6M dataset. Each mean value is represented by a dot, and the associated standard deviation is shown by the bars, indicating the variability around the mean bone lengths.}
\label{fig:bone distribution}
\end{figure}

Before introducing our augmentation method, we briefly discuss the approach used in \cite{chen2021anatomy}. They randomly adjust bone lengths $L$ based on the average bone lengths in the training batch, denoted by $\bar{L} = \left[ \bar{l}'_1 \cdots \bar{l}'_{J-1} \right]^T \in \mathbb{R}^{(J-1) \times 1}$.

\begin{equation}
    l'_i = l_i + r_i \bar{l}_i,\quad r_i \sim \mathcal{U}(-0.3,0.3),\quad i=1,\dots,J-1
    \label{eq:random aug}
\end{equation}

The proportion varies between $-30\%$ to $30\%$, which can lead to $L'$ deviating from natural human anatomical structures. For example, this method could generate an unnaturally long forearm combined with a short upper arm. Additionally, $L'$ might lack symmetry because each bone is adjusted independently by different random values. In our experiments, we ensure symmetry by applying identical random adjustments to corresponding bones on both sides of the body.

As shown in Figure \ref{fig:bone distribution}, the variability of each bone length is different. For instance, subjects in the Human3.6M dataset \cite{h36m_pami,IonescuSminchisescu11} have similar lengths of forearms but differ in lengths of upper arms. Intuitively, we may randomly adjust the bone lengths by applying a normal distribution:

\begin{equation}
    \begin{split}
        l'_i &\sim \mathcal{N}(l_i, \sigma_i),\quad i=1,\dots,J-1 \\
        l'_i &= l'_j,\quad \text{if they are the same body part on different side.}
    \end{split}
    \label{eq:random aug std}
\end{equation}
where the mean value is the $i$-th original length and $\sigma_i$ denotes the standard deviation of the $i$-th bone length in the Human3.6M dataset. We also maintain the symmetry in this case.

\subsubsection{Synthetic Bone Lengths.}
SMPL \cite{Loper2015SMPL} is a model that generates 3D human meshes from parameters. We use SMPL to randomly generate human meshes and then evaluate the bone lengths from these meshes. This ensures that the bone lengths are symmetric and reasonable, reflecting natural body shapes. To evaluate bone lengths from meshes, we apply the joint regression matrix $\mathcal{J}$ introduced in \cite{choi2020pose2mesh} to mesh coordinates $M$:

\begin{equation}
    \tilde{L} = \mathcal{J}M
    \label{eq:joint regression}
\end{equation}

The 3D poses in the Human3.6M dataset are recorded using a marker-based motion capture system, where the position of each joint depends on the placement of the markers. Consequently, a single joint regression matrix cannot accurately describe the positions of the joints. When using a single joint regression matrix, the distribution of the regressed bone lengths differs from that of the Human3.6M dataset, as shown in Figure \ref{fig:bone distribution}, leading to poor predictive ability. To mitigate the difference in data distribution, we align the mean value of the regressed bone lengths with the mean value in the Human3.6M dataset. After the alignment, we obtain the augmented bone lengths $L'$.

\subsection{Bone Length Model}\label{sect:length model}

\begin{figure}
    \centering
    \begin{subfigure}[b]{0.45\textwidth}
        \includegraphics[height=48mm]{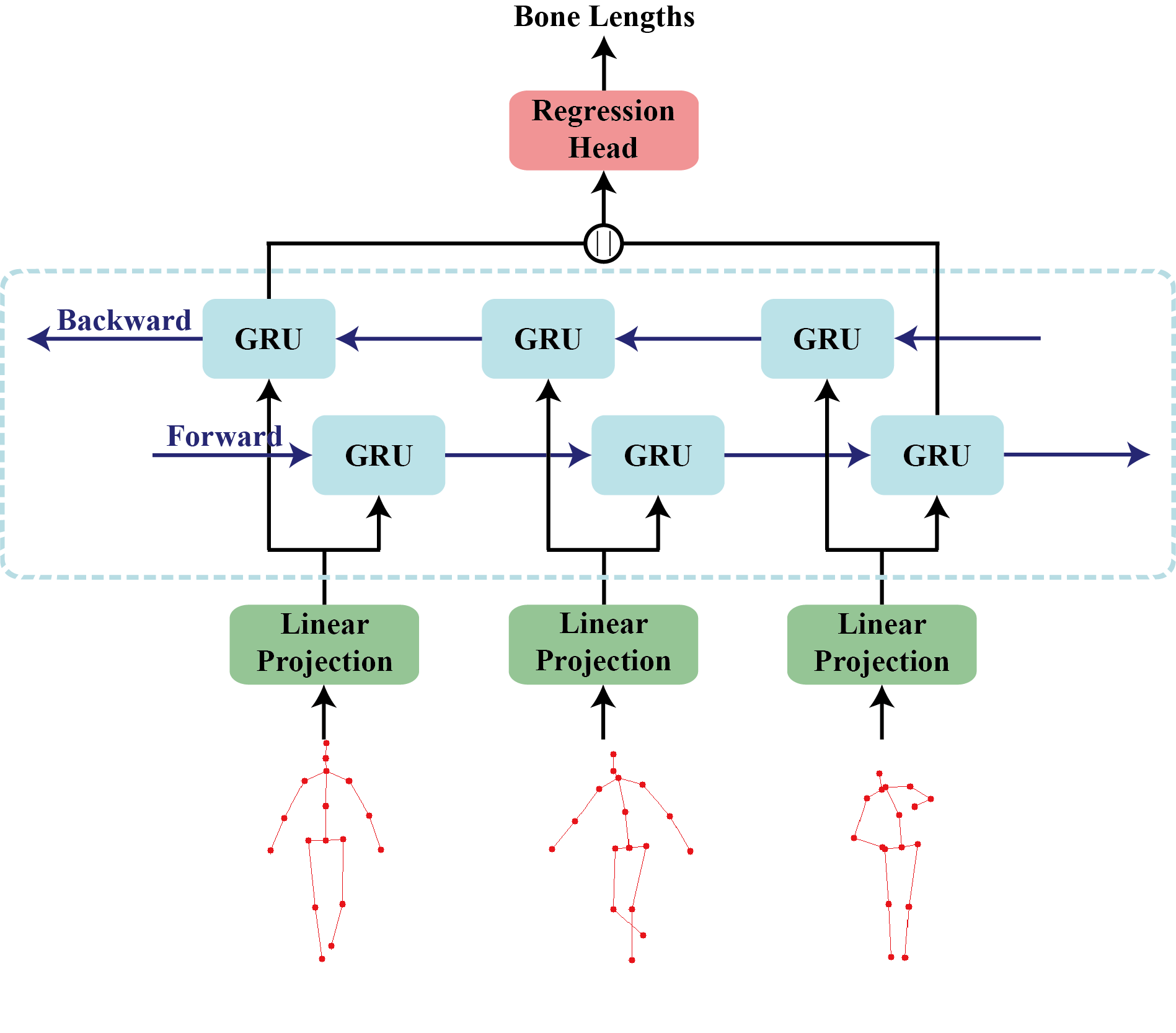}
        \caption{Bi-GRU model}
        \label{fig:length biGRU model}
    \end{subfigure}
    \hfill
    \begin{subfigure}[b]{0.45\textwidth}
        \includegraphics[height=48mm]{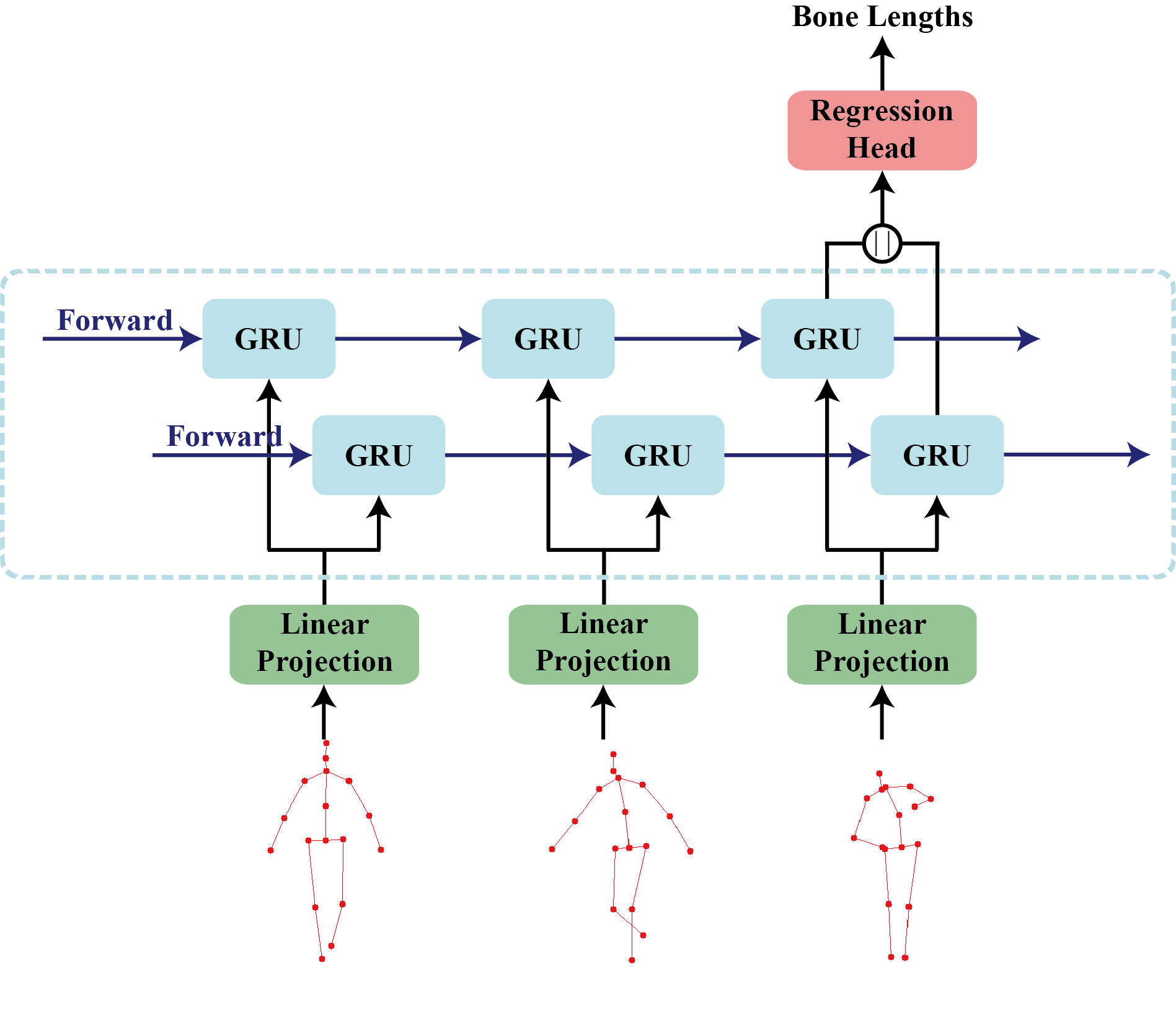}
        \caption{GRU model}
        \label{fig:length GRU model}
    \end{subfigure}
    
    \caption{The structures of our bone length prediction models. The input length is 3 for illustration.}
    \label{fig:length model}
\end{figure}

The structures of our models are illustrated in Figure \ref{fig:length model}. Our primary model, depicted in Figure \ref{fig:length biGRU model}, utilizes a single-layer bidirectional gated recurrent unit (Bi-GRU) \cite{chung2014empirical}. This Bi-GRU model processes the entire sequence of 2D keypoints from a given video, leveraging both past and future information for improved prediction accuracy. However, due to its reliance on future data, the Bi-GRU model is not suitable for real-time online processing. To address this limitation, we also developed a GRU model, shown in Figure \ref{fig:length GRU model}, which updates bone lengths by processing the input keypoints frame by frame, making it suitable for online applications. In this section, we specifically introduce the Bi-GRU model.

During training, we slice the sequences of 2D keypoints into fixed-size segments for convenience. The input sequence of 2D keypoints is denoted by $X = \left[ \bm{x}_0 \cdots \bm{x}_N \right] \in \mathbb{R}^{N \times (J \times 2)}$, where $N$ is the sequence length, $J$ is the number of joints, and $\bm{x}_t \in \mathbb{R}^{2J}$ is the flattened vector of 2D keypoints at frame $t$. A linear projection layer maps each $\bm{x}_t$ to a higher dimension $c$.
\begin{equation}
    \bm{x}'_t = W_p \bm{x}_t + \bm{b}_p.
    \label{eq:linear projection}
\end{equation}
where $W_p$ is the weight matrix and $\bm{b}_p$ is the bias vector in the linear projection layer.

The projected vectors $\bm{x}'_t \in \mathbb{R}^c$ are then input to the GRU. The forward process at frame $t$ can be written as
\begin{equation}
    \bm{h}_t = \text{GRU}(X'_t,\ \bm{h}_{t-1})
    \label{eq:GRU}
\end{equation}
where $\bm{h}_t \in \mathbb{R}^{c'}$ is the hidden state at frame $t$ with hidden size $c'$, and the initial hidden state $\bm{h}_0$ is a zero vector. The backward process is similar but processes $X'_t$ in reverse order. We concatenate the final hidden states from the forward process and backward processes to obtain $\bm{h} \in \mathbb{R}^{2c'}$. The bone lengths $L \in \mathbb{R}^{(J-1) \times 1}$ are then regressed from $\bm{h}$ using the weight matrix $W_R$.
\begin{equation}
    L = W_R \bm{h}
    \label{eq:regress head}
\end{equation}

Our goal is to minimize the difference between the predicted bone lengths $L$ and the groundtruth bone lengths $\hat{L}$. The loss function is defined by the mean absolute error:
\begin{equation}
    \mathcal{L}_L = \frac{1}{J-1} \sum\limits_{i=1}^{J-1} \Vert l_i - \hat{l}_i \Vert_1
    \label{eq:bone length loss}
\end{equation}

\subsection{Bone Length Adjustment and Fine-tuning}\label{sect:length adjustment}

\begin{figure}
\centering
\includegraphics[height=35mm]{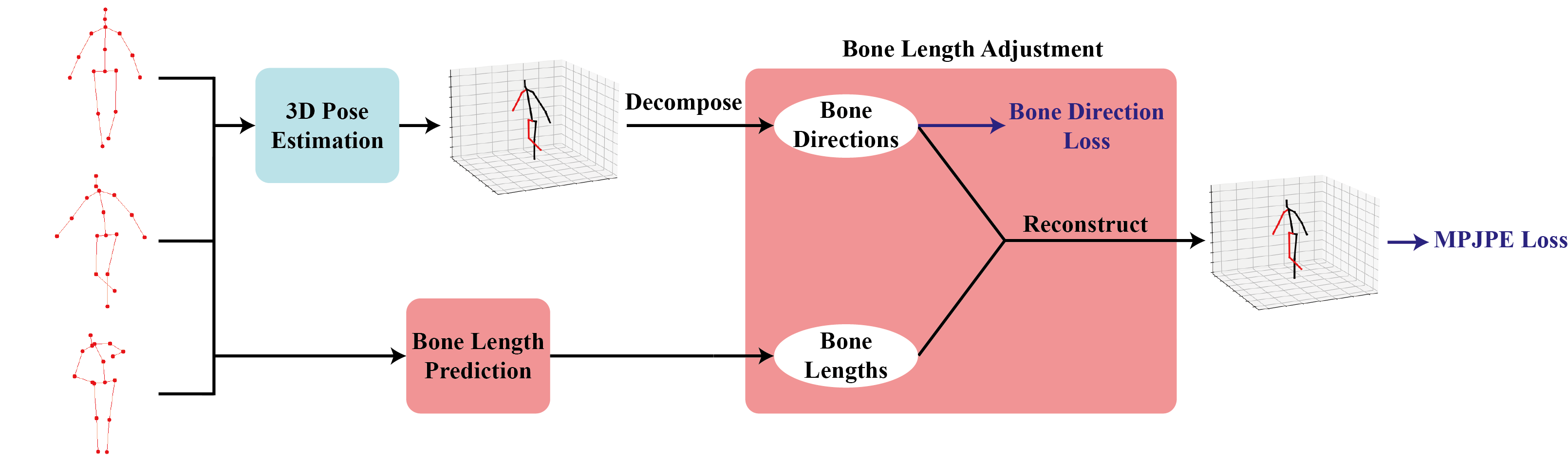}
\caption{The overview of bone length adjustment. The 3D pose estimation is based on existing 2D-to-3D lifting models. The blue part is based on existing lifting models. Only the parameters in blue part are fine-tuned.}
\label{fig:adjustment}
\end{figure}

Figure \ref{fig:adjustment} provides an overview of our bone length adjustment method. This technique is applied to the human poses predicted by existing 2D-to-3D lifting models. The bone length adjustment involves replacing the bone lengths of the human poses with our predicted bone lengths, as illustrated in Figure \ref{fig:replacement}.

Given a sequence of 2D keypoints $X$ and a lifting model, we first obtain predicted poses $P$ from the lifting model. The sequence $X$ is segmented to fit in the input requirements of the lifting model. We then decompose the poses $P$ into bone lengths $L$ and bone directions $D$. Concurrently, we use the entire sequence $X$ to predict new bone lengths $L'$ with our model. By combining the bone directions $D$ from the lifting model and the bone lengths $L'$ from our model, we generate the reconstructed poses $P'$. This process refines the poses, ensuring a more realistic body structure.

To evaluate the adjustment process, we use the Mean Per Joint Position Error (MPJPE) to measure the error between the reconstructed pose $P'$ and the groundtruth pose $\hat{P}$:

\begin{equation}
    \mathcal{L}_P = \frac{1}{J} \sum\limits_{i=0}^{J-1} \Vert \bm{p}'_i - \hat{\bm{p}}_i \Vert_2
    \label{eq:MPJPE}
\end{equation}

\subsubsection{Fine-tuning.}
In our adjustment process, bone lengths can also enhance the lifting models' ability to predict bone directions. We propose a fine-tuning method based on our adjustment process. The bone length prediction model is fixed during this process to prevent overfitting since the data without augmentation is used. We fix the weights of our bone length model and fine-tune the lifting models by minimizing the error in the predicted bone directions and the MPJPE of the reconstructed pose $P'$. The direction loss is defined as:

\begin{equation}
    \mathcal{L}_D = \frac{1}{J-1} \sum\limits_{i=1}^{J-1} \Vert \bm{d}_i - \hat{\bm{d}}_i \Vert_2
    \label{eq:bone direction loss}
\end{equation}
where $\bm{d}_i$ is the predicted direction and $\hat{\bm{d}}_i$ is the groundtruth direction of the $i$-th bone. The total loss combines both the direction loss and the position error loss:

\begin{equation}
    \mathcal{L} = \mathcal{L}_D + \mathcal{L}_P
    \label{eq:total loss}
\end{equation}

\section{Experimental setup}

\subsection{Dataset and Evaluation}

We evaluate our method using the Human3.6M dataset \cite{h36m_pami,IonescuSminchisescu11}, which contains 3.6 million frames of 11 actors performing 15 diverse actions, recorded by four synchronized cameras at 50 Hz. Seven subjects are annotated with 3D poses. Following the standard protocol in prior works \cite{pavllo20193d:videopose,chen2021anatomy,zheng20213d:poseformer,li2022mhformer,zhang2022mixste}, we train our model on five subjects (S1, S5, S6, S7, S8) and test on two subjects (S9, S11), using a 17-joint skeleton. For bone length evaluation, we use the bone length error as described by Equation \ref{eq:bone length loss}, comparing the predicted to the groundtruth lengths. For human poses, we use two protocols: Protocol 1 measures the Mean Per Joint Position Error (MPJPE), the average Euclidean distance between the predicted and groundtruth joint positions, and Protocol 2 (P-MPJPE) reports the error after aligning the predicted poses with the groundtruth in terms of translation, rotation, and scale, enabling us to better evaluate the accuracy of the predicted body structure.

\subsection{Implementation Details}

\subsubsection{Bone Length Prediction.}
We align the mean value of our synthetic bone lengths with the Human3.6M dataset. Our study evaluates five different methods for generating bone lengths, detailed in Section \ref{sect:length aug}:

\begin{itemize}
    \item \textbf{Random Uniform Distribution}: Randomly generated bone lengths from a uniform distribution.
    \item \textbf{Random Normal Distribution}: Bone lengths generated from a normal distribution using standard deviations in the training set.
    \item \textbf{Random Normal Distribution (Involving Test Data STD)}: Bone lengths generated from a normal distribution using standard deviations in the entire dataset, involving test data statistics.
    \item \textbf{Synthetic}: Synthetic bone lengths aligned with the mean values in the training set.
    \item \textbf{Synthetic (Involving Test Data Mean Values)}: Synthetic bone lengths aligned with the mean values in the entire dataset, involving test data statistics.
\end{itemize}

During training, we use a sequence length $N=512$ and utilize the entire sequence during testing. The projected dimension $c$ is set to $256$, and the hidden state dimension $c'$ is set to $512$. We train our models using the Adam optimizer with an exponentially decaying learning rate schedule. The initial learning rate is set to $0.0001$, and it decays by a factor of $0.95$ each epoch. The batch size is set to 
$256$. We train our model with groundtruth 2D keypoints which are projected from synthetic poses and test our model with 2D keypoints predicted by the Cascaded Pyramid Network (CPN) \cite{Chen2018cascaded:CPN}.

\subsubsection{Fine-tuning.}
For fine-tuning, we select Pavllo \etal \cite{pavllo20193d:videopose} as our fine-tuning target. We configure the sequence length $N$ to $243$, and apply horizontal flip augmentation during both training and testing, following their settings. We fine-tune the model using the Adam optimizer and a batch-normalization momentum set to the final state $0.001$. Similarly, we employ an exponentially decaying learning rate schedule, starting at $0.00004$ with a decay factor of $0.95$ per epoch. The batch size for fine-tuning is set to $1024$, consistent with their work. We utilize 2D keypoints predicted by CPN for both training and testing phases. Finally, the horizontal flip augmentation is applied at train and test time, following previous works \cite{pavllo20193d:videopose,chen2021anatomy,zheng20213d:poseformer}.

\section{Results}

In this section, we compare our results to recent works. Since our method involves adjusting model predictions, we reproduced the results by following the instructions provided by each work. This may result in slight variations due to differences in environments.

For diffusion models, they generate multiple hypotheses and select the best one to evaluate the error. As outlined in \cite{shan2023diffusion}, there are four methods to select the best hypothesis. The reported errors in these models are based on selecting the joint-wise hypothesis closest to the ground truth, which is impractical in real-world applications. Instead, we applied a joint-level aggregation method to generate the final prediction without relying on the ground truth. Therefore, the errors reported here will differ from those claimed in the original works.

\subsection{Bone Length Model} \label{sect:length model result}

\begin{figure}[t]
    \centering
    \includegraphics[height=44mm]{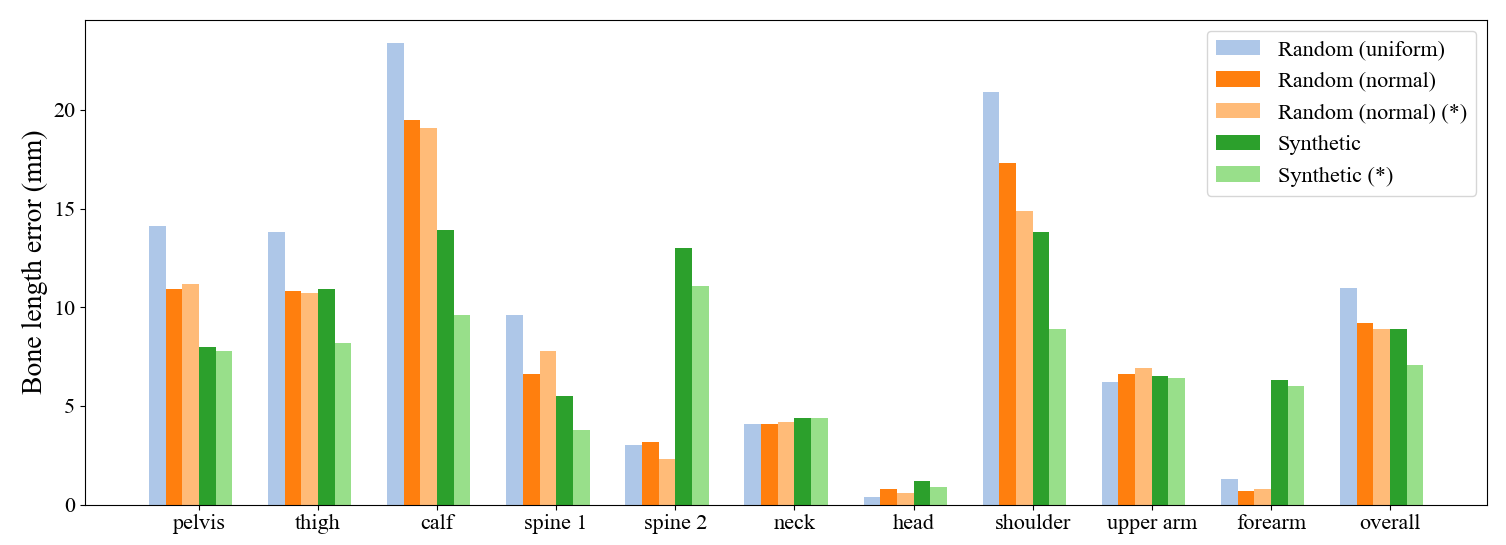}
    \caption{The average bone length error comparison across all frames of the test set in Human3.6M. ($*$) including test data statistics.}
    \label{fig:bone error}
\end{figure}

\begin{table}[t]
    \centering
    \caption{Quantitative comparison of bone length error. Best in bold and second best underlined. $(*)$ including test data statistics. $(\dag)$ diffusion model. $(-)$ bone length model only.}
    \label{tbl:bone error}
    
    \begin{tabular}{|ll|c|c|c|c|}
    \hline
    \multicolumn{2}{|l|}{} & Bone length error ↓ (mm) \\
    \hline
    Pavllo \etal \cite{pavllo20193d:videopose} & CVPR'19 & 12.3 \\[2pt]
    Chen \etal \cite{chen2021anatomy} (-) & TCSVT'21 & 10.3 \\[2pt]
    Chen \etal \cite{chen2021anatomy} & TCSVT'21 & 8.9 \\[2pt]
    Zheng \etal \cite{zheng20213d:poseformer}& ICCV'21 & 10.8 \\[2pt]
    Gong \etal \cite{gong2023diffpose} (\dag) & CVPR'23 & \underline{8.5} \\[2pt]
    Shan \etal \cite{shan2023diffusion} (\dag) & ICCV'23 & 10.6 \\[2pt]
    Peng \etal \cite{peng2024ktpformer} (\dag) & CVPR'24 & 10.9 \\[1pt]
    \hline
    Ours, Bi-GRU (synthetic) (-) & & 8.9 \\[2pt]
    Ours, Bi-GRU (synthetic) (*)(-) & & \textbf{7.1} \\[2pt]
    Ours, GRU (synthetic) (*)(-) & & \textbf{7.1} \\[1pt]
    \hline
    \end{tabular}
\end{table}

Figure \ref{fig:bone error} presents the outcomes of our Bi-GRU bone length model evaluation. Utilizing synthetic bone lengths during training time augmentation yields the lowest overall bone length error among all methods evaluated. The random uniform distribution method fail to generate bone lengths that adhere to natural human body proportions, resulting in poor performance. Conversely, synthetic methods demonstrate superior performance over random methods, even when not using test data statistics.

Table \ref{tbl:bone error} shows the comparison of bone lengths. For the lifting models, the error is evaluated by decomposing the predicted poses into bone lengths. Our model achieves the state-of-the-art result when not using test data statistics. Additionally, the GRU model designed for online processing performs comparably to the Bi-GRU model. The synthetic method incorporating test data statistics notably outperforms all other results. Given that the training set comprises data from only five subjects, the statistics may not fully represent broader variations, leading to significant disparities between using and not using test data statistics. With a more comprehensive dataset, our approach could potentially circumvent such limitations.

\begin{figure}[t]
    \centering
    \includegraphics[height=50mm]{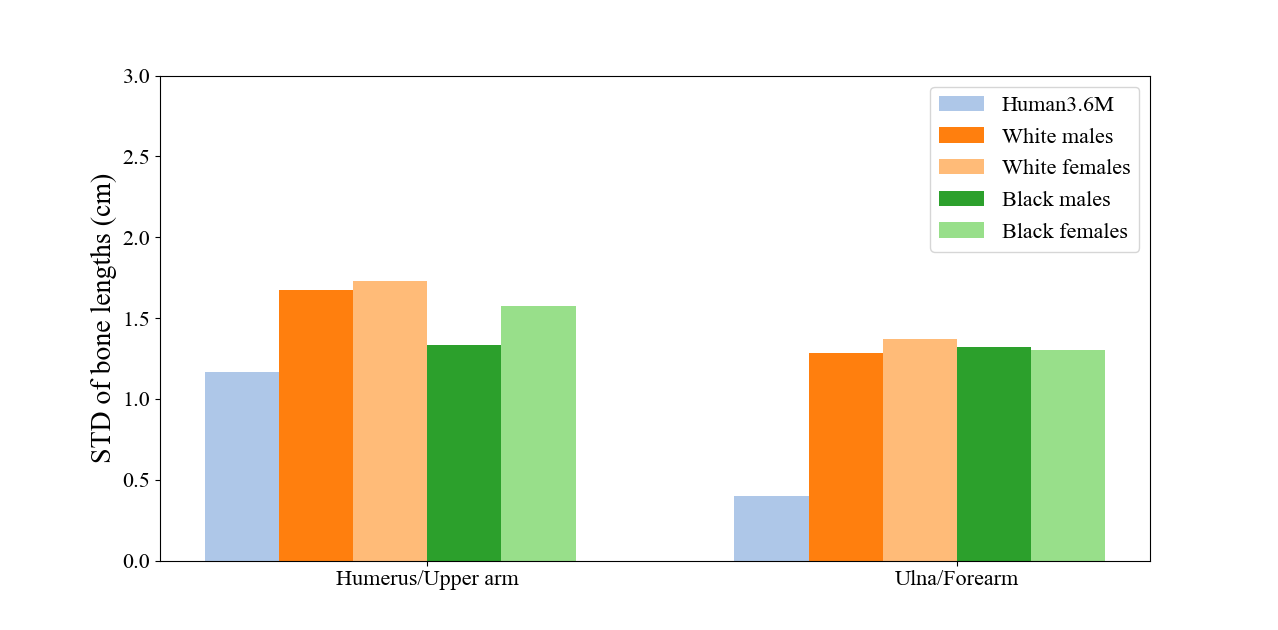}
    \caption{Comparison between the standard deviation of real bone lengths and bone lengths in Human3.6M.}
    \label{fig:STD bone length}
\end{figure}

Examining Figure \ref{fig:bone distribution}, we observe that the standard deviations of certain bones, such as spine 2, neck, head, and forearm in the Human3.6M dataset are exceptionally small. As shown in Figure \ref{fig:STD bone length}, anthropometric research \cite{trotter1952estimation} indicates similar standard deviations for lengths of the humerus (upper arm) and ulna (forearm). However, in Human3.6M, standard deviations for the lengths of the upper arm and forearm differ significantly. Additionally, the standard deviation of the upper arm closely matches the anthropometric result for the humerus, while the standard deviation of the lower arm is much smaller than the anthropometric result for the ulna. These discrepancies may arise from limitations in the transformation process from MoCap raw data to human poses, potentially influenced by constraints on body shape or inaccuracies in marker placement within the Human3.6M dataset.

Even with dataset constraints, our synthetic method consistently outperforms random methods and the lifting models. We select the model trained with synthetic bone lengths using statistics of the test data as our final model.

\subsection{Adjustment and Fine-tuning}

\begin{table}
    \scriptsize
    \centering
    \caption{Quantitative comparison of the adjustment process on reconstruction error evaluated on Human3.6M under MPJPE and P-MPJPE. Best results of the same base model are in bold.}
    \label{tbl:adjusment result}
    
    \begin{tabular}{|ll|c|c|c|c|}
    \hline
    \multicolumn{2}{|l|}{Base model} & \makecell{Bone length error \\ ↓ (mm)} & Bone length model & \makecell{MPJPE \\ ↓ (mm)} & \makecell{P-MPJPE \\ ↓ (mm)} \\
    \hline
    \multirow{3}{*}{Pavllo \etal \cite{pavllo20193d:videopose}}
            & \multirow{3}{*}{CVPR'19} & \multirow{3}{*}{12.3}
            & \xmark & 46.8 & 36.5 \\
            & & & GRU & 45.6 & 36.1 \\
            & & & Bi-GRU & \textbf{45.2} & \textbf{35.8} \\
    \hline
    \multirow{3}{*}{Chen \etal \cite{chen2021anatomy}}
            & \multirow{3}{*}{TCSVT'21} & \multirow{3}{*}{8.9}
            & \xmark & 44.2 & 35.0 \\
            & & & GRU & 44.0 & 34.8 \\
            & & & Bi-GRU & \textbf{43.5} & \textbf{34.5} \\
    \hline
    \multirow{3}{*}{Zheng \etal \cite{zheng20213d:poseformer}}
            & \multirow{3}{*}{ICCV'21} & \multirow{3}{*}{10.8}
            & \xmark & 44.3 & 34.6 \\
            & & & GRU & 43.3 & 34.1 \\
            & & & Bi-GRU & \textbf{42.9} & \textbf{33.8} \\
    \hline
    \multirow{3}{*}{Li \etal \cite{li2022mhformer}}
            & \multirow{3}{*}{CVPR'22} & \multirow{3}{*}{10.3}
            & \xmark & 43.0 & 34.5 \\
            & & & GRU & 42.5 & 34.0 \\
            & & & Bi-GRU & \textbf{42.2} & \textbf{33.7} \\
    \hline
    \multirow{3}{*}{Zhang \etal \cite{zhang2022mixste}}
            & \multirow{3}{*}{CVPR'22} & \multirow{3}{*}{11.0}
            & \xmark & 40.9 & 32.7 \\
            & & & GRU & 40.6 & 32.5 \\
            & & & Bi-GRU & \textbf{40.2} & \textbf{32.2} \\
    \hline
    
    \multirow{3}{*}{Gong \etal \cite{gong2023diffpose}}
            & \multirow{3}{*}{CVPR'23} & \multirow{3}{*}{8.5}
            & \xmark & 39.5 & 31.2 \\
            & & & GRU & 39.4 & 31.1 \\
            & & & Bi-GRU & \textbf{39.0} & \textbf{30.8} \\
    \hline
    \multirow{3}{*}{Tang \etal \cite{tang20233d}}
            & \multirow{3}{*}{CVPR'23} & \multirow{3}{*}{10.7}
            & \xmark & 41.8 & 32.9 \\
            & & & GRU & 41.6 & 33.1 \\
            & & & Bi-GRU & \textbf{41.0} & \textbf{32.7} \\
    \hline
    \multirow{3}{*}{Zhao \etal \cite{zhao2023poseformerv2}}
            & \multirow{3}{*}{CVPR'23} & \multirow{3}{*}{11.1}
            & \xmark & 45.2 & 35.6 \\
            & & & GRU & 44.5 & 35.2 \\
            & & & Bi-GRU & \textbf{44.0} & \textbf{34.9} \\
    \hline
    \multirow{3}{*}{Shan \etal \cite{shan2023diffusion}}
            & \multirow{3}{*}{ICCV'23} & \multirow{3}{*}{10.6}
            & \xmark & 39.6 & 31.7 \\
            & & & GRU & 38.9 & 31.2 \\
            & & & Bi-GRU & \textbf{38.6} & \textbf{31.0} \\
    \hline
    \multirow{3}{*}{Peng \etal \cite{peng2024ktpformer}}
            & \multirow{3}{*}{CVPR'24} & \multirow{3}{*}{10.9}
            & \xmark & 40.2 & 32.2 \\
            & & & GRU & 39.9 & 31.9 \\
            & & & Bi-GRU & \textbf{39.4} & \textbf{31.5} \\
    \hline
    \multirow{3}{*}{Xu \etal \cite{xu2024finepose}}
            & \multirow{3}{*}{CVPR'24} & \multirow{3}{*}{12.2}
            & \xmark & 40.2 & 32.9 \\
            & & & GRU & 40.1 & 32.5 \\
            & & & Bi-GRU & \textbf{39.6} & \textbf{32.2} \\
    \hline
    \end{tabular}
\end{table}

Table \ref{tbl:adjusment result} illustrates the reconstruction error of existing lifting models before and after applying our adjustment method. Across all tested models, we observe consistent performance improvements under both protocol 1 (MPJPE) and protocol 2 (P-MPJPE) after adjustment with both GRU and Bi-GRU models. Our Bi-GRU model outperforms the GRU model, showing the advantage of utilizing future information. The degree of improvement correlates with the bone length error inherent in the original models. Models with larger initial bone length errors, such as \cite{pavllo20193d:videopose}, demonstrate significant enhancement, achieving a 3\% reduction in Protocol 1 error with the adjustment. In contrast, models like \cite{chen2021anatomy}, which exhibit smaller bone length errors due to effective bone length prediction, show more modest improvements of around 1\% under Protocol 1. Our adjustment effectively rectifies pose errors for all models under Protocol 2, which undergoes rigid alignment like scaling. This indicates that our predicted bone lengths possess better body proportions.

\begin{table}
    \fontsize{5}{7}\selectfont
    
    \centering
    \caption{Reconstruction error on Human3.6M before and after adjustment and fine-tuning with our Bi-GRU model. The top table shows the result under protocol 1. The bottom table shows the result under protocol 2. Best in bold and second best underlined. Unit: millimeter}
    \label{tbl:finetune result}
    
    \resizebox{\columnwidth}{!}{
    \begin{tabular}{|l|ccccccccccccccc|c|}
    \hline
    &&&&&&&&&&&&&&&&\\[-1em]
    Protocol 1 & Dir. & Disc. & Eat & Greet & Phone & Photo & Pose & Purch. & Sit & SitD. & Smoke & Wait & WalkD. & Walk & WalkT. & Avg \\
    &&&&&&&&&&&&&&&&\\[-1em]
    \hline
    &&&&&&&&&&&&&&&&\\[-1em]
    Pavllo \etal \cite{pavllo20193d:videopose} & 45.2 & 46.7 & 43.3 & 45.6 & 48.1 & 55.1 & 44.6 & 44.3 & 57.3 & 65.8 & 47.1 & 44.0 & 49.0 & 32.8 & 33.9 & 46.8 \\[1pt]
    Adjustment & 41.9 & \textbf{45.2} & 42.1 & 44.3 & 46.1 & \textbf{53.5} & 43.4 & \textbf{42.1} & 55.5 & 64.0 & 45.5 & 42.3 & 47.1 & 32.1 & 32.9 & 45.2 \\[1pt]
    Fine-tuning & \textbf{41.5} & \textbf{45.2} & \textbf{41.9} & \textbf{44.0} & \textbf{45.9} & 53.6 & \textbf{43.3} & 42.3 & \textbf{55.3} & \textbf{63.8} & \textbf{45.3} & \textbf{42.2} & \textbf{47.0} & \textbf{31.8} & \textbf{32.4} & \textbf{45.0} \\[1pt]
    
    \hline
    \multicolumn{17}{c}{} \\
    \hline
    &&&&&&&&&&&&&&&&\\[-1em]
    Protocol 2 & Dir. & Disc. & Eat & Greet & Phone & Photo & Pose & Purch. & Sit & SitD. & Smoke & Wait & WalkD. & Walk & WalkT. & Avg \\
    &&&&&&&&&&&&&&&&\\[-1em]
    \hline
    &&&&&&&&&&&&&&&&\\[-1em]
    Pavllo \etal \cite{pavllo20193d:videopose} & 34.1 & 36.1 & 34.4 & 37.2 & 36.4 & 42.2 & 34.4 & 33.6 & 45.0 & 52.5 & 37.4 & 33.8 & 37.8 & 25.6 & 27.3 & 36.5 \\[1pt]
    Adjustment & 32.8 & \textbf{35.1} & \textbf{33.1} & 36.2 & \textbf{35.2} & 42.0 & 33.8 & \textbf{33.1} & \textbf{44.1} & \textbf{52.0} & 36.5 & \textbf{33.0} & \textbf{37.3} & 25.0 & 27.0 & 35.8 \\[1pt]
    Fine-tuning & \textbf{32.7} & 35.3 & \textbf{33.1} & \textbf{35.9} & 35.3 & \textbf{41.7} & \textbf{33.7} & 33.3 & \textbf{44.1} & \textbf{52.0} & \textbf{36.4} & \textbf{33.0} & \textbf{37.3} & \textbf{24.9} & \textbf{26.6} & \textbf{35.7} \\[1pt]
    \hline
    \multicolumn{17}{c}{} \\
    \end{tabular}
    }
\end{table}

Table \ref{tbl:finetune result} details the results of our fine-tuning process. We select Pavllo \etal \cite{pavllo20193d:videopose} as the target lifting model and fine-tune it. We focus on comparing the result after adjustment and the result after fine-tuning. While the lifting model already performs well, fine-tuning demonstrates incremental improvements, particularly noticeable in dynamic actions like "Walk" (0.3 mm) and "Walk Together" (0.5 mm). This highlights the effectiveness of leveraging bone length cues to refine the model's predictions.

\begin{table}
    \scriptsize
    \centering
    \caption{Comparison on Parameters, frame per second (FPS), and MPJPE. The evaluation is performed without test-time augmentation.}
    \label{tbl:FPS test}
    
    \begin{tabular}{|l|c|c|c|c|}
    \hline
    &&&&\\[-4pt]
    Model & Frames & Parameters (M) & FPS & MPJPE (mm) \\
    &&&&\\[-4pt]
    \hline
    &&&&\\[-4pt]
    Pavllo \etal \cite{pavllo20193d:videopose} & 243 & 16.95 & 958 & 46.8 \\[2pt]
    Chen \etal \cite{chen2021anatomy} & 243 & 59.18 & 197 & 44.2 \\[2pt]
    Zheng \etal \cite{zheng20213d:poseformer} & 81 & 9.60 & 379 & 44.3 \\[2pt]
    \hline
    &&&&\\[-4pt]
    Pavllo \etal \cite{pavllo20193d:videopose} (with adjustment) & 243 & 19.34 & 435 & 45.6 \\[2pt]
    Chen \etal \cite{chen2021anatomy} (with adjustment) & 243 & 61.57 & 154 & 44.0 \\[2pt]
    Zheng \etal \cite{zheng20213d:poseformer} (with adjustment) & 81 & 11.99 & 252 & 43.3 \\[2pt]
    \hline
    &&&&\\[-4pt]
    Chen \etal \cite{chen2021anatomy} (bone length model) & - & 8.56 & 715 & - \\[2pt]
    Ours, GRU model & - & 2.39 & 2097 & - \\[2pt]
    \hline
    \end{tabular}
\end{table}

In Table \ref{tbl:FPS test}, we evaluate the inference efficiency across three scenarios: (1) the lifting models alone, (2) the lifting models with our adjustment process, and (3) the bone length models.

We measure the frames per second (FPS) for these models during real-time online processing, where each model predicts a single frame at a time. The horizontal flip augmentation is not applied in the evaluation. We repeat the inference step 10,000 times, simulating a test on a 10,000-frame video, using a single GeForce GTX 3060 Ti GPU. As our Bi-GRU model is unsuitable for online processing, we test using our GRU model instead. After applying our adjustment process, the MPJPE loss improves significantly with minimal overhead in model size and computation time. Although \cite{pavllo20193d:videopose} with adjustment runs at half the FPS compared to without adjustment, it remains faster than the other models listed. Additionally, the complete human pose estimation includes 2D keypoint detection and 2D-to-3D lifting. The FPS of most 2D keypoint detection models is lower than 100. Thus our approach will not be the bottleneck.

For the bone length model, our approach updates bone length values faster than \cite{chen2021anatomy}. Our model requires only the input of the new frame at each step, as past information is stored in the hidden state, whereas \cite{chen2021anatomy} needs to randomly select 50 frames from previous inputs to predict bone lengths. The FPS is limited by our adjustment process that we decompose poses into bone directions and reconstruct poses with inferred bone lengths.

In summary, our adjustment and fine-tuning methodologies enhance the robustness and accuracy of existing 3D lifting models, demonstrating their efficacy in improving pose estimation across different evaluation protocols and dynamic scenarios. In real-time online processing, our adjustments achieve competitive results with minimal efficiency overhead.

\section{Conclusion}

We introduced a novel approach to enhance 3D human pose estimation by integrating bone length prediction and adjustment methods. Using a single-layer bidirectional GRU with synthetic data augmentation, our bone length model achieved the lowest error on the Human3.6M dataset. The bone length adjustment technique significantly reduced MPJPE and P-MPJPE errors in 2D-to-3D lifting models. Fine-tuning further improved pose accuracy, particularly for dynamic actions. Additionally, our GRU model operates efficiently in real-time, demonstrating minimal overhead from the adjustment process.

Overall, our approach effectively enhances the anatomical accuracy of 3D human pose predictions, demonstrating significant improvements in error metrics and robustness across various models.

%
%
\bibliographystyle{splncs04}
\bibliography{main}
\end{document}